\documentclass[11pt]{amsart} 
\usepackage[utf8]{inputenc}
\usepackage{graphicx}
\usepackage{hyperref}
\usepackage{amssymb}
\usepackage[table]{xcolor}
\usepackage{booktabs,multirow}

\begin{document}

\title[SynapCountJ]{SynapCountJ --- a Tool for Analyzing Synaptic Densities in Neurons}

\author[G. Mata et al.]{Gadea Mata \and Jónathan Heras \and Miguel Morales \and Ana Romero \and Julio Rubio}
\address{\small \rm  Gadea Mata, Jónathan Heras, Ana Romero, and Julio Rubio\\
Departamento de Matemáticas y Computación. Universidad de La Rioja.}
\email{\{gadea.mata,jonathan.heras,ana.romero,julio.rubio\}@unirioja.es}

\address{\small \rm  Miguel Morales\\
Instituto de Neurociencias, Universidad Autónoma de Barcelona, Barcelona, Spain}
\email{miguelmorales@spineup.es}


\maketitle

\begin{abstract}
The quantification of synapses is instrumental to measure the evolution of synaptic densities of neurons under the effect of some physiological conditions, neuronal diseases or even drug treatments. However, the manual quantification of synapses is a tedious, error-prone, time-consuming and subjective task; therefore, tools that might automate this process are desirable. In this paper, we present SynapCountJ, an ImageJ plugin, that can measure synaptic density of individual neurons obtained by immunofluorescence techniques, and also can be applied for batch processing of neurons that have been obtained in the same experiment or using the same setting. The procedure to quantify synapses implemented in SynapCountJ is based on the colocalization of three images of the same neuron (the neuron marked with two antibody markers and the structure of the neuron) and is inspired by methods coming from Computational Algebraic Topology. SynapCountJ provides a procedure to semi-automatically quantify the number of 
synapses of neuron cultures; as a result, the time required for such an analysis is greatly reduced. 

\end{abstract}




\section{Introduction}

Synapses are the points of connection between neurons, and they are dynamic structures subject to a continuous process of formation and elimination. Pathological conditions, such as the Alzheimer disease, have been related to synapse loss associated with memory impairments. Hence, the possibility of changing the number of synapses may be an important asset to treat neurological diseases~\cite{selkoe02}. To this aim, it is necessary to determine the evolution of synaptic densities of neurons under the effect of some physiological conditions, neuronal diseases or even drug treatments.  

The procedure to quantify synaptic density of a neuron is usually based on the colocalization between the signals generated by two antibodies~\cite{CEC11}. Namely, neuron cultures are permeabilized and treated with two different primary markers (for instance, bassoon and synapsin). These antibodies recognize specifically two presynaptic structures. Then, it is necessary a secondary antibody couple attached to different fluorochromes (for instance red and green; note, that several other combinations of color are possible) making these two synaptic proteins visible under the fluorescence microscope. The two markers are photographed in two gray-scale images; that, in turn, are overlapped using respectively the red and green channels. In the resultant image, the yellow points (colocalization of the code channels) are the candidates to be the synapses. 

The final step in the above procedure is the selection of the yellow points that are localized either on the dendrites of the neuron or adjacent to them. Tools like MetaMorph~\cite{metamorph} or ImageJ~\cite{imagej} --- a Java platform for image processing that can be easily extended by means of plugins --- can be used to manually count the number of synapses; however, such a manual quantification is a tedious, time-consuming, error-prone, and subjective task; hence, tools that might automate this process are desirable. In this paper, we present \emph{SynapCountJ}, an ImageJ plugin, that semi-automatically quantifies synapses and synaptic densities in neuron cultures. 

\section{Methodology}

SynapCountJ supports two execution modes: individual treatment of a neuron and batch processing. 

\subsection{Individual treatment of a neuron}

\begin{figure}
\centering
\includegraphics[scale=0.05]{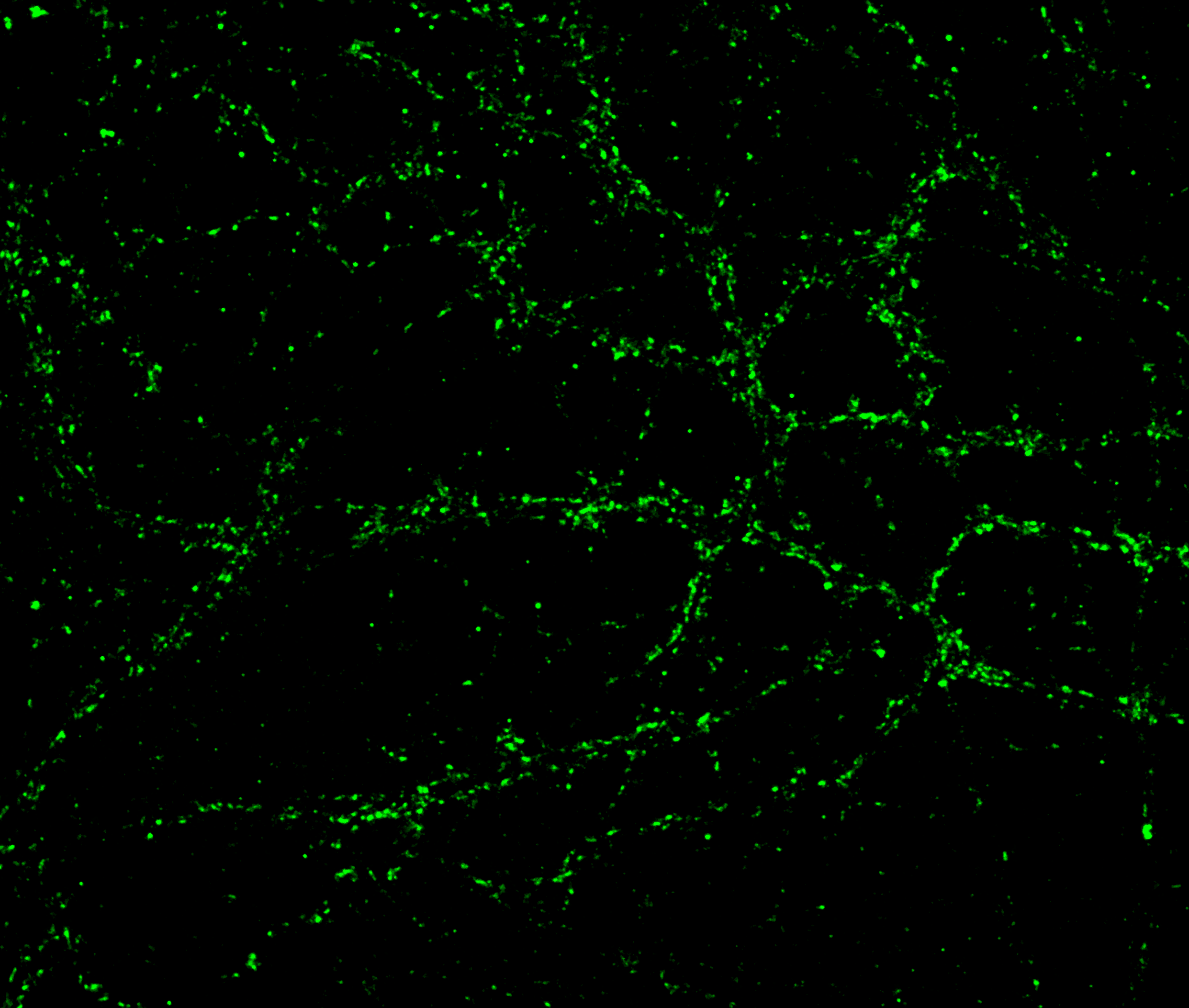}
\includegraphics[scale=0.05]{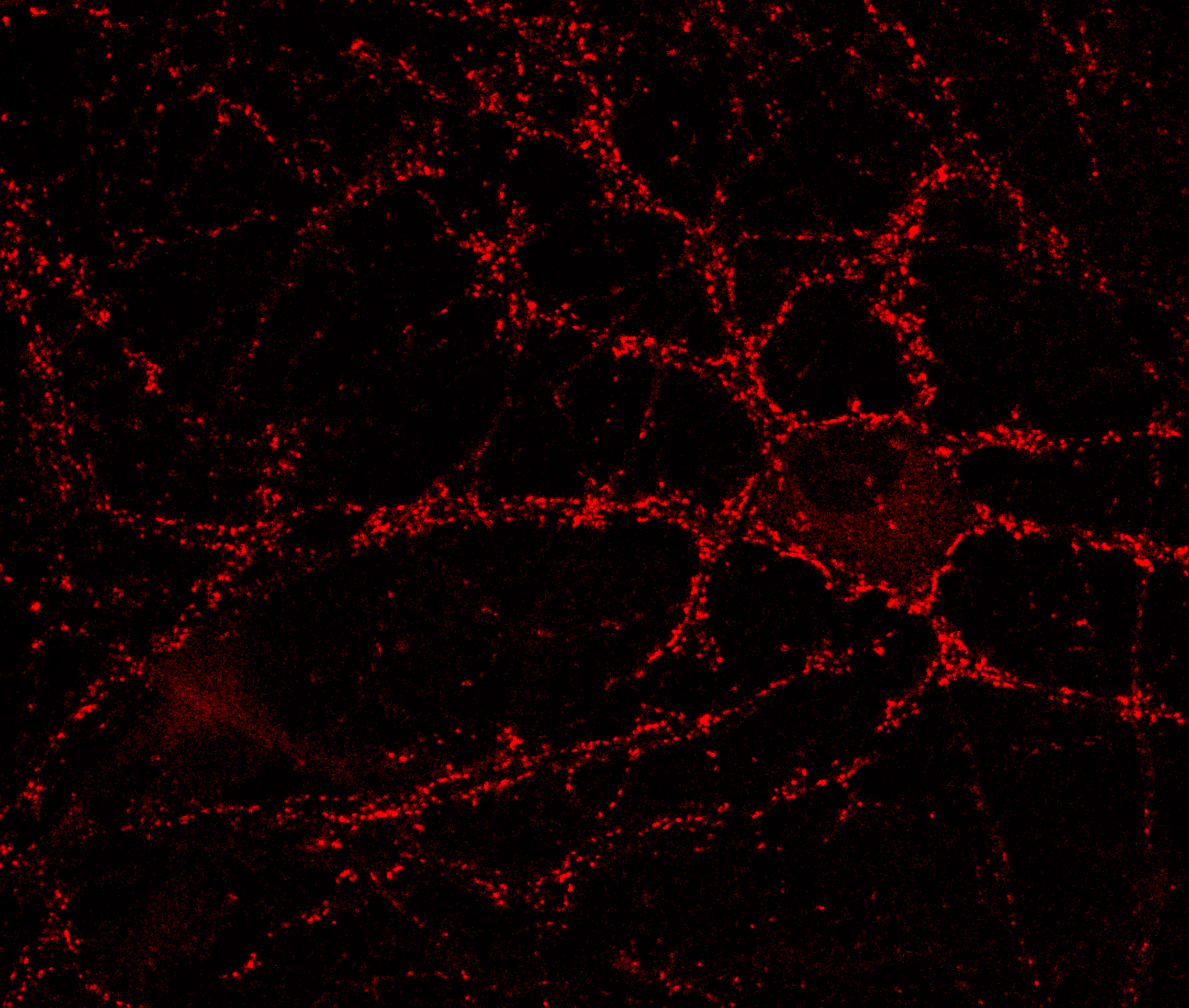}
\includegraphics[scale=0.05]{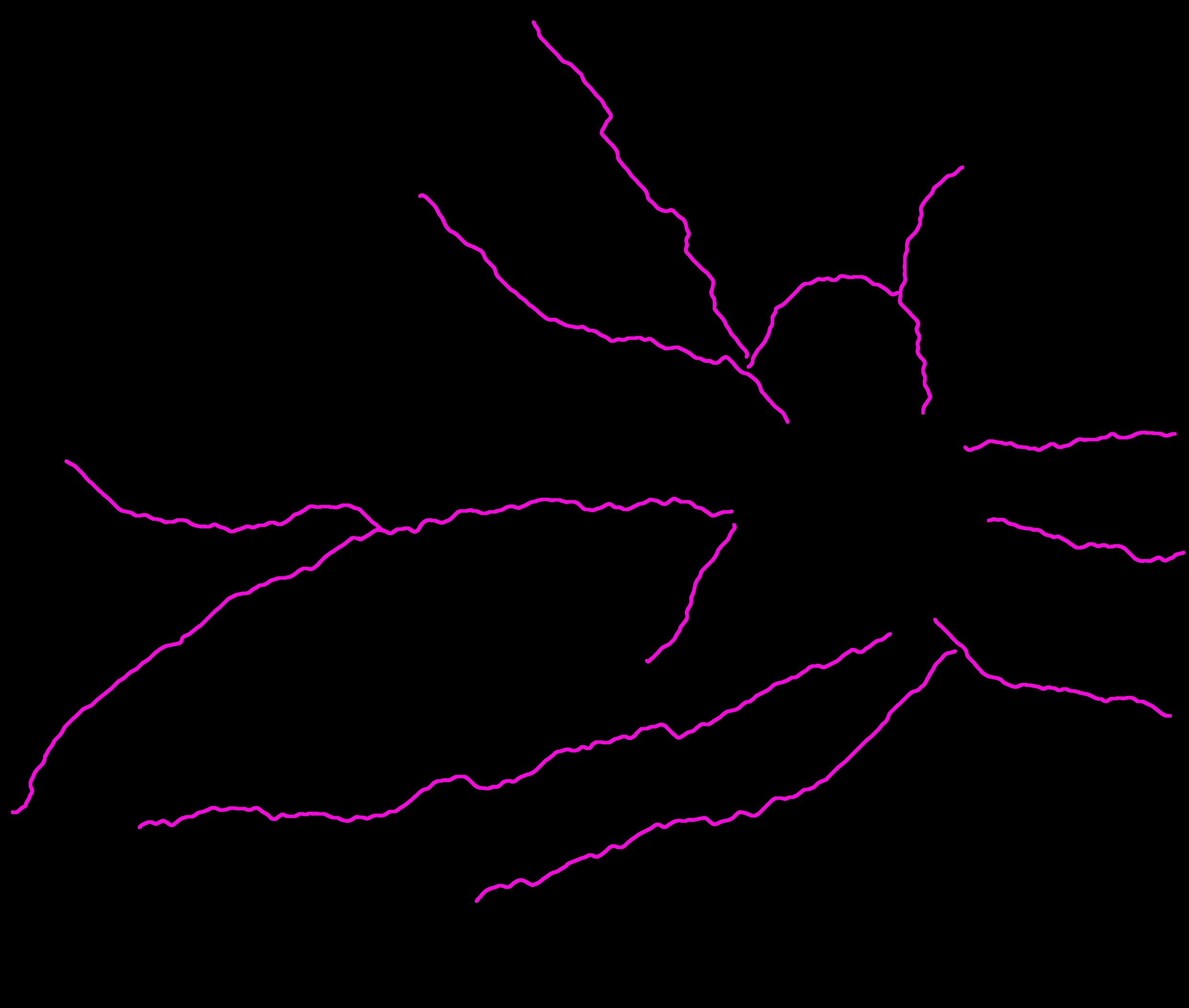}
\caption{\textbf{Neuron with two antibody markers and its structure}. \emph{Left.} Neuron marked with the bassoon antibody marker. \emph{Center.} Neuron marked with the synapsin antibody marker. \emph{Right.} Structure of the neuron.}\label{fig:firststeps}
\end{figure}

The input of SynapCountJ in this execution mode are two images of a neuron marked with two antibodies (an image per antibody), see Figure~\ref{fig:firststeps}. SynapCountJ is able to read tiff (a standard format for biological images) and lif files (obtained from Leica confocal microscopes) --- the latter requires the Bio-Formats plugin~\cite{bioformats}. The following steps are applied to quantify the number of synapses in the given images. 

In the first step, from one of the two images, the region of interest (i.e. the dendrites where the quantification of synapses will be performed) is specified using NeuronJ~\cite{neuronj}  --- an ImageJ plugin for tracing elongated image structures. In this way, the background of the image is removed. The result is a file containing the traces of each dendrite of the image.
 
\begin{figure}
\centering
  \includegraphics[scale=0.3]{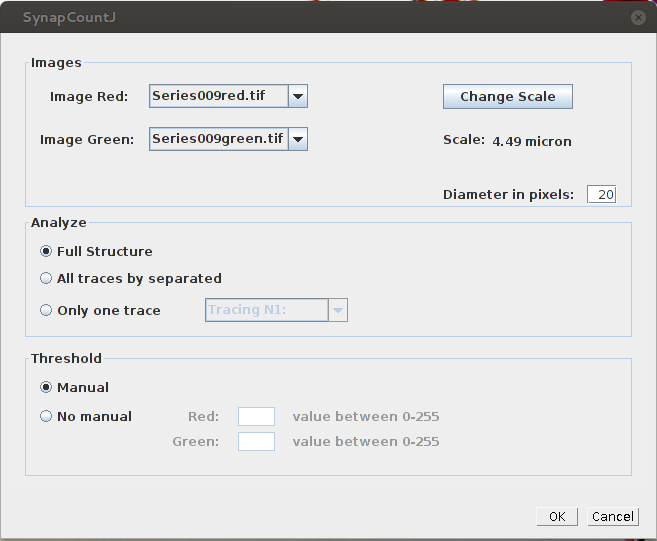}
\caption{\textbf{SynapCountJ window to configure the analysis}.}\label{fig:configuration}
\end{figure}

Subsequently, the user can decide whether she wants to perform a global analysis of the whole neuron, or a local analysis focused on each dendrite of the neuron. In both cases, SynapCountJ requires additional information  such as the scale and the mean thickness (that is determined by the size of the subjacent dendrite) of the region to analyze (see Figure~\ref{fig:configuration}) --- these parameters determine the area of the dendrite avoiding the background (i.e. all the non-synaptic marking).  

\begin{figure}
\centering
 \includegraphics[scale=0.5]{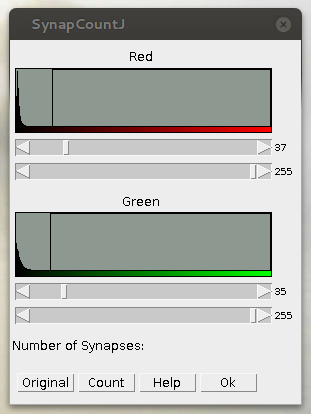}
 \includegraphics[scale=0.2]{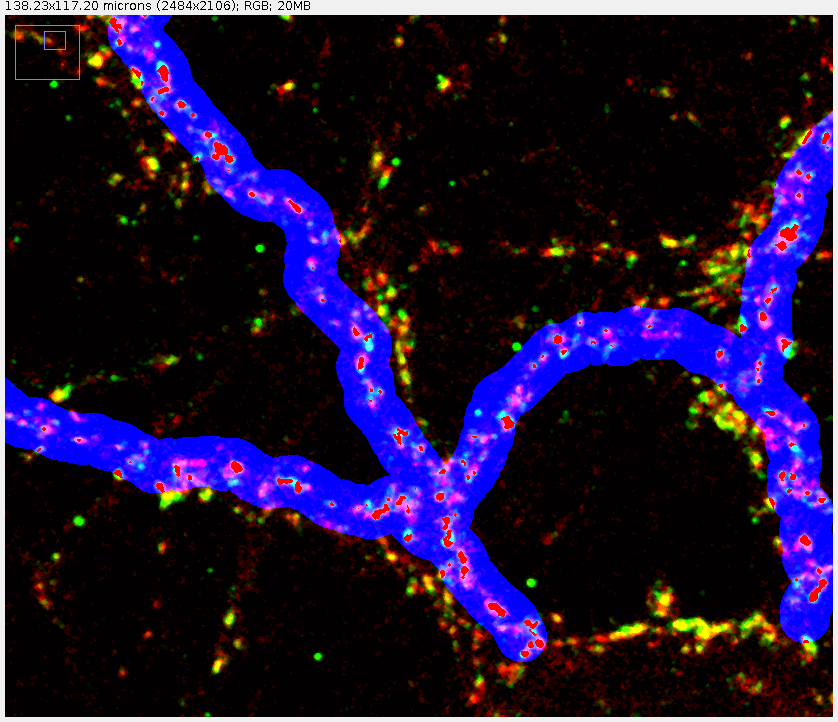}
\caption{\textbf{SynapCountJ window to modify the threshold of the red and green channels}. \emph{Left.} Window to fix the threshold of the image. \emph{Right.} Fragment of the neuron image with the synapses indicated as the red areas on the structure of the neuron marked in blue.  Moving the scrollbars of left window, the marked areas of the image are changed.}\label{fig:threshold}
\end{figure}

Taking into account the settings provided by the user, SynapCountJ overlaps the two original images of the neuron and the structure of the neuron previously defined. From the resultant image, SynapCountJ identifies the almost white points (the result of green, red, and blue combination) as synaptic candidates, and it allows the user to modify the values of the red and green channels in order to modify the detection threshold (see Figure~\ref{fig:threshold}). 

Once that the detection threshold has been fixed, the counting process is started. Such a process is inspired by techniques coming Computational Algebraic Topology. In spite of being an abstract mathematical subject, Algebraic Topology has been successfully applied in digital image analysis~\cite{SGF03,GDR05,MMRR15}. In our particular case, the white areas are segmented from the overlapped image, and the colors of the resultant image are inverted --- obtaining as a result a black-and-white image where the synapses are the black areas. From such an image, the problem of quantifying the number of synapses is reduced to compute the homology group in dimension $0$ of the image; this corresponds to the computation of the number of connected components of the image. 

\begin{figure}
\centering
\includegraphics[scale=0.5]{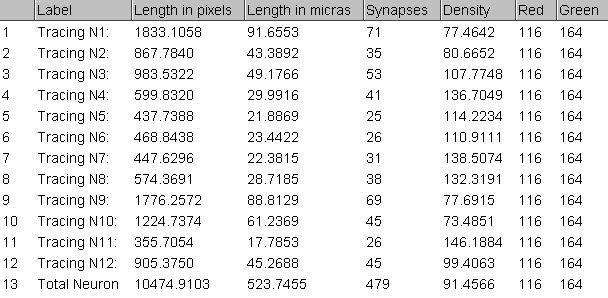}

\includegraphics[scale=0.35]{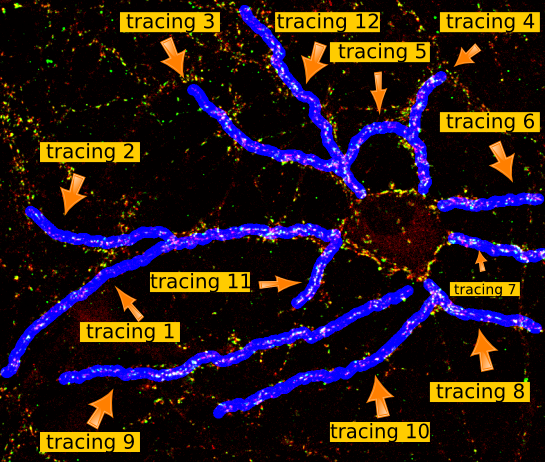}
\includegraphics[scale=0.18]{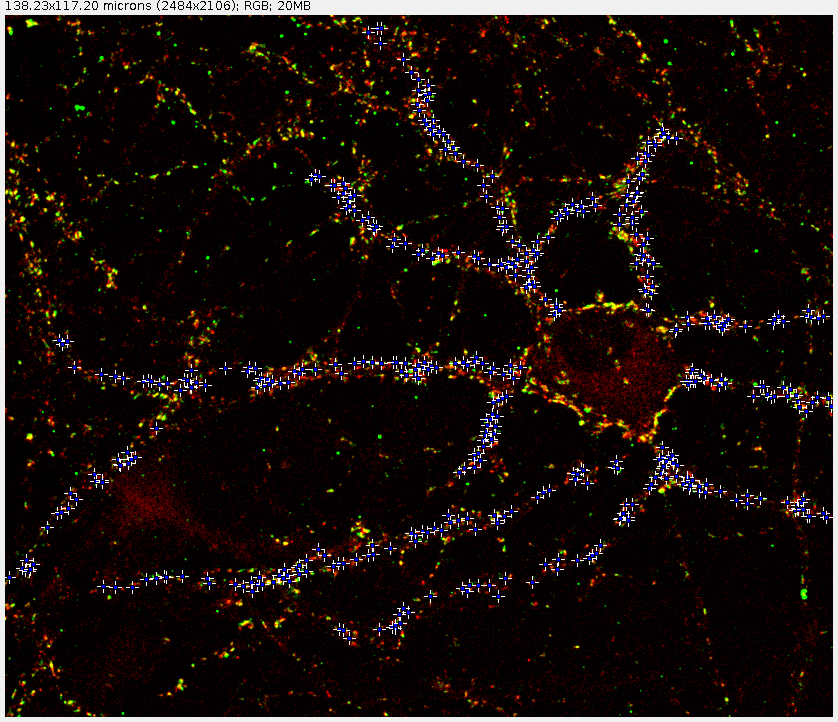}
\caption{\textbf{Results provided by SynapCountJ}. \emph{Top}. Table with the results obtained by SynapCountJ. \emph{Bottom Left}. Image with the analyzed region of the neuron. \emph{Bottom Right}. Image with the counted synapses indicated by means of blue crosses.}\label{fig:results}
\end{figure}

Finally, SynapCountJ returns a table with the obtained data (length of dendrites both in pixels and micras, number of synapses, and density of synapses per 100 micron) and two images showing, respectively, the analyzed region and the marked synapses (see Figure~\ref{fig:results}).

\subsection{Batch processing}\label{subsec:batch}

Images obtained from the same biological experiment usually have similar settings; hence, their processing in SynapCountJ will use the same configuration parameters. In order to deal with this situation, SynapCountJ can be applied for batch processing of several images using a configuration file generated from the analysis of an individual neuron.  

For batch processing, SynapCountJ reads tiff files organized in folders or a lif file (the kind of files produced by Leica confocal microscopes), and using the configuration file processes the different images. As a result, a table with the information related to each neuron from the batch (the table includes an analysis for both the whole neuron and from each of its dendrites) is obtained. 

\section{Experimental Results}

The original aim of SynapCountJ was the automatic analysis of synaptic density on neurons treated with SB 415286 --- an organic inhibitor of GSK3, a kinase which inhibition was proposed as a therapy in AD treatment~\cite{DSC03} --- such a treatment, as it was previously demonstrated, promotes synaptogenesis and spinogenesis in primary cultures of rodent hippocampal neurons and in Drosophyla neurons~\cite{CJA15,FBB04}. In this setting, a comparative study has been performed in order to evaluate the results that can be obtained with SynapCountJ. 

Primary hippocampal cultures were obtained from P0 rat pups (Sprague-Dawley, strain, Harlan Laboratories Models SL, France). Animals were anesthetized by hypothermia in paper-lined towel over crushed-ice surface during 2-4 minutes and euthanized by decapitation. Animals were handled and maintained in accordance with the Council Directive guidelines 2010/63EU of the European Parliament. Primary cultures of hippocampus neurons were prepared as previously described in~\cite{MCG00}. Briefly, glass coverslips (12 mm in diameter) were coated with poly-L-lysine and laminin, 100 and 4 $\mu$g/ml respectively. Neurons at a $10\times104$ neurons/cm$^2$ density were seeded and grown in Neurobasal (Invitrogen, USA) culture medium supplemented with glutamine 0.5 mM, 50 mg/ml penicillin, 50 units/ml streptomycin, 4\% FBS and 4\% B27 (Invitrogen, CA, USA), as described before in~\cite{CEC11}. At days 4, 7 and 14 in culture a 20\% of culture medium was replace by fresh medium. Cytosine-D-arabinofuranoside (4 $\mu$M) was 
added to prevent overgrowth of glial cells (day 4).

Synaptic density on hippocampal cultures was identified as previously described in~\cite{CEC11}. In short, cultures were rinsed in phosphate buffer saline (PBS) and fixed for 30 min in 4\% paraformaldehyde-PBS. Coverslips were incubated overnight in blocking solution with the following antibodies: anti-Bassoon monoclonal mouse antibody (ref. VAM-PS003, Stress Gen, USA) and rabbit polyclonal sera against Synapsin (ref. 2312, Cell Signaling, USA). Samples were incubated with a fluorescence-conjugated secondary antibody in PBS for 30 min. After that, coverslips were washed three times in PBS and mounted using Mowiol (all secondary antibodies from Molecular Probes-Invitrogen, USA). Stack images (pixel size 90 nm with 0.5 $\mu$m Z step) were obtained with a Leica SP5 Confocal microscope (40x lens, 1. 3 NA). Percentage of synaptic change is the average of different cultures under the same experimental conditions. As a control, we used sister untreated cultures growing in the same 24 well multi plate.

\begin{figure}
\centering
\includegraphics[scale=.6]{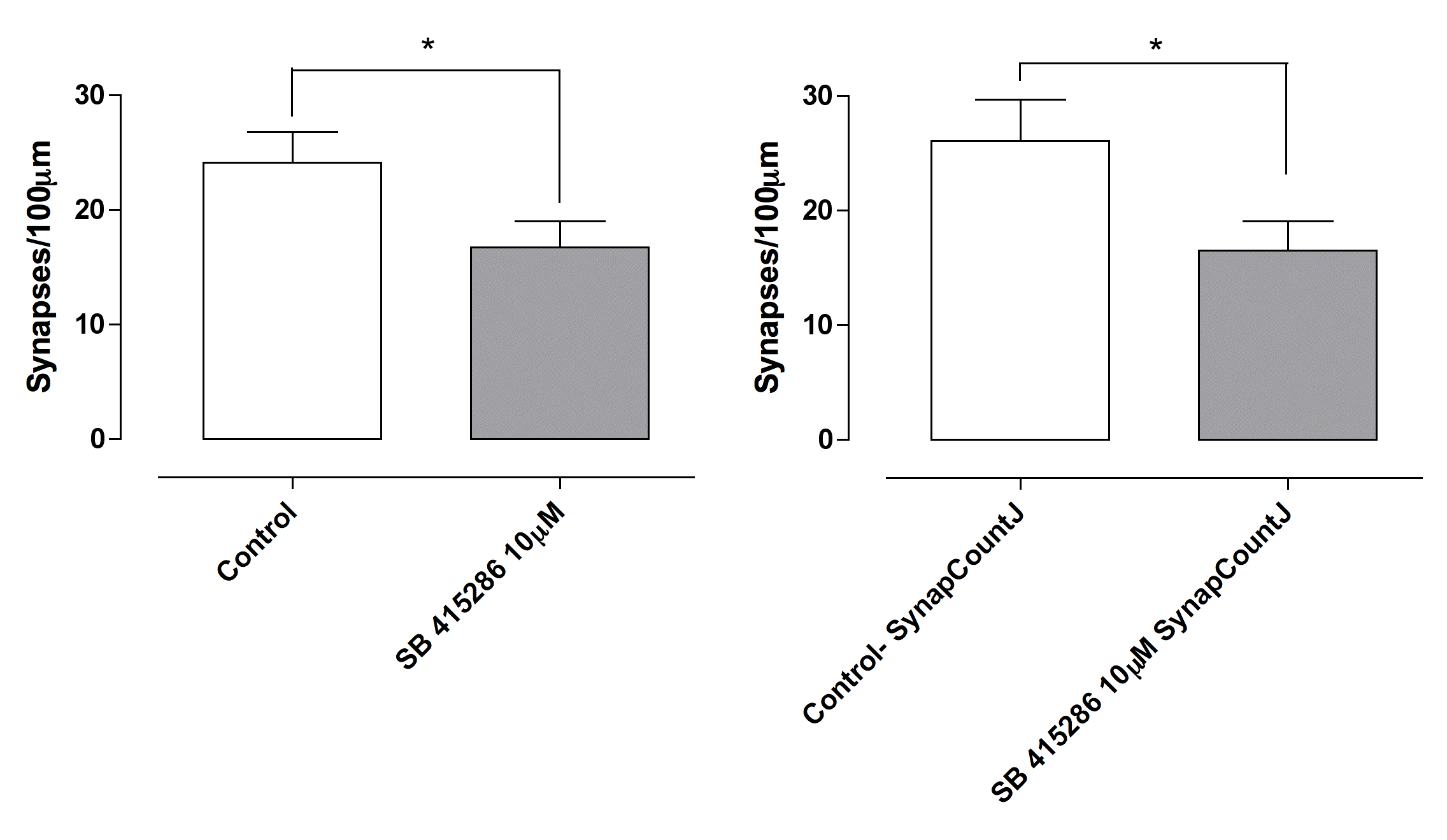}
\caption{\textbf{Quantification of synapses}. \emph{Left}. Manual quantification of synapses. \emph{Right}. Quantification of synapses using \emph{SynapCountJ}.}\label{fig:comparison}
\end{figure}

A total of 13 individual images from three independent cultures has been analyzed. In Figure~\ref{fig:comparison} we can observe that using a manual method to identify and count synapses, we obtain a mean of $24.12$ synapses in control cultures and $16.74$ in treated cultures. The results obtained with SynapCountJ are similar, there is a mean of $26.03$ synapses in control cultures and $16.50$ in the ones which have been treated. 

Notwithstanding the differences in the quantification, in both procedures we obtain almost the same inhibition percentage, a 30.51\% manually and 36.61\% automatically. This shows the suitability of SynapCountJ to count synapses, meaning a considerably reduction of the time employed in the
manual process. Namely, the manual analysis of an image takes approximately 5 minutes; of a batch, 1 hour; and, of a complete study, 4 hours. Using SynapCountJ, the time to analyze an image is 30 seconds; a batch, 2 minutes; and, a complete study, 6 minutes. 

\section{Discussion}

Up to the best of our knowledge, 4 tools have been developed to quantify synapses and measure synaptic density: Green and Red Puncta~\cite{greenandred}, Puncta Analyzer~\cite{punctaanalyzer}, SynD~\cite{synd} and SynPAnal~\cite{synpanal} --- a summary of the general features of these tools can be seen in Table~\ref{general-features}. The rest of this section is devoted to compare SynapCountJ with these tools, the comparison is summarized in Table~\ref{analysis-features}. 

\begin{table}
\centering
{\scriptsize
\rowcolors{1}{white}{black!10!white}
\begin{tabular}{ccp{2cm}p{1.5cm}p{2cm}}
\toprule
Software & Language & Underlying Technology & Types of Images & Technique for detection\\
\midrule
Green and Red Puncta & Java & ImageJ & tiff & Colocalization\\
Puncta Analyzer  & Java & ImageJ2 & tiff & Colocalization\\
SynapCountJ & Java & ImageJ & tiff and lif & Colocalization\\
SynD & Matlab & Matlab & tiff and lsm & Brightness\\
SynPAnal & Java &  & tiff & Brightness\\
\bottomrule
\end{tabular}}
\caption{\textbf{General features of the analyzed software}.}\label{general-features}
\end{table}

There are two approaches to locate synapses in an RGB image either based on  colocalization or brightness. In the former, synapses are identified as the colocalization of bright points in the red and green channels --- this is the approach followed by  Green and Red Puncta, Puncta Analyzer and SynapCountJ --- in the latter, synapses are the bright points of a region of an image --- the approach employed in SynD and SynPAnal. In both approaches, it is necessary a threshold that can be manually adjusted to increase (or decrease) the number of detected synapses; such a functionality is supported by all the tools.  

In the quantification of synapses from RGB images, it is instrumental to determine the region of interest (i.e. the dendrites of the neurons where the synapses are located); otherwise, the analysis will not be precise due to noise coming from irrelevant regions or the background of the image --- this happens in the Green and Red Puncta tool since it considers the whole image for the analysis. Puncta Analyzer allows the user to fix a rectangle containing the dendrites of the neuron, but this is not completely precise since some regions of the rectangle might contain points considered as synapses that do not belong to the structure of the neuron. SynD is the only software that automatically detects the dendrites of a neuron; however, it can only be applied to neurons with a cell-fill marker, and does not support the analysis from specific regions, such as soma or distal dendrites. SynapCountJ and SynPAnal provide the functionality to manually draw the dendrites of the image; allowing the user to 
designate the specific areas where quantification is restricted.  

The main output produced by all the available tools is the number of synapses of a given image; additionally, SynapCountJ, SynD and SynPAnal provides the length of the dendrites; and, SynapCountJ is the only tool that outputs the synaptic density per micron. All the tools but Green and Red Punctua can export the results to an external file for storage and further processing. 

Finally, as we have explained in Subsection~\ref{subsec:batch}, images obtained from the same biological experiment usually have similar settings; hence, batch processing might be useful. This functionality is featured by SynapCountJ and SynD, and requires a previous step of saving the configuration of an individual analysis. SynPAnal does not support batch processing, but the configuration of an individual analysis can be saved to be later applied in other individual analysis. 

\begin{table}
\centering
{\scriptsize
\rowcolors{1}{white}{black!10!white}
\begin{tabular}{p{1.75cm}p{1.75cm}cp{1.25cm}p{1.5cm}ccc}
\toprule
Software & Detection of dendrites & Threshold & Batch Processing & Dendrites length & Density & Export & Save\\
\midrule
Green and Red Puncta & Not used & \checkmark &  & & & & \\ 
Puncta Analyzer      & Manual ROI & \checkmark &  & & & \checkmark & \\
SynapCountJ          & Manual & \checkmark & \checkmark  & \checkmark & \checkmark & \checkmark & \checkmark\\
SynD                 & Automatic & \checkmark & \checkmark & \checkmark &  & \checkmark & \checkmark\\
SynPAnal             & Manual & \checkmark&  & \checkmark &  & \checkmark & \checkmark\\
\bottomrule
\end{tabular}}
\caption{\textbf{Features to quantify synapses and synaptic density of the analyzed software}.}\label{analysis-features}
\end{table}

As a summary, SynapCountJ is more complete than the rest of available programs. It can use different types of synaptic markers and can process batch images. Furthermore, a differential feature of SynapCountJ is that it is based on a topological algorithm (namely, computing the number of connected components in a combinatorial structure), allowing us to validate the correctness of our approach by means of formal methods in software engineering~\cite{PDH14}. 

\section{Conclusions and Further Work}

SynapCountJ is an ImageJ plugin that provides a semi-automatic procedure to quantify synapses and measure synaptic density from immunofluorescence images obtained from neuron cultures. This plugin has been tested not only with neurons in development, but also with the neuromuscular union of Drosophila; therefore, it can be applied to the study of images that contain two synaptic markers and a determined structure. The results obtained with SynapCountJ are consistent with the results obtained manually; and SynapCountJ dramatically reduces the time required for the quantification of synapses. 

As further work, it remains the tasks of improving the usability of the plugin and including post-processing tools to manually edit the obtained results. Additionally, and since the final aim of our project is the complete automation of the whole process, it is necessary a procedure to automatically detect the neuron morphology. 

\section*{Availability and Software Requirements}

SynapCountJ is an ImageJ plugin that can be downloaded, together with its documentation, from  \url{http://imagejdocu.tudor.lu/doku.php?id=plugin:utilities:synapsescountj:start}. 
SynapCountJ is open source and available for use under the GNU General Public License. This plugin runs within 
both ImageJ and Fiji~\cite{fiji} and has been tested on Windows, Macintosh and Linux machines. \\

\section*{Acknowledgements}
This work was supported by the Ministerio de Econom\'ia y Competitividad projects [MTM2013-41775-P, MTM2014-54151-P, BFU2010-17537]. G. Mata was also supported by a PhD grant awarded by the University of La Rioja [FPI-UR-13]. 

\noindent \emph{Conflict of Interest:} none declared.

\bibliographystyle{plain} 
\bibliography{biblio}
\end{document}